\documentclass[letterpaper,10pt,conference]{ieeeconf}

\usepackage{graphicx}
\usepackage[dvipsnames]{xcolor}
\usepackage{url}
\usepackage{amsmath,amssymb}
\usepackage[caption=false,font=footnotesize]{subfig}
\usepackage{stfloats}
\usepackage{hyperref}
\usepackage{siunitx}

\IEEEoverridecommandlockouts
\title{\LARGE \bf
Fly, Drive, Reconfigure: A Modular Reconfigurable Aerial-Ground Platform for Field Operations
}

\author{Li-Yu Lo$^{1*}$, Yanbaihui Liu$^{1*}$, Chengchuan Shu$^{1}$, Tyler Harris$^{1}$, Jonathan Ryan$^{1}$ and Boyuan Chen$^{1}$ \\
\vspace{-0.4cm}
\textcolor{orange}{\href{http://generalroboticslab.com/HARP}{www.generalroboticslab.com/HARP}} 
\thanks{*Equal Contribution. $^{1}$ All authors are from Duke University.}%
\thanks{This work is supported by a Dean’s Research Venture Fund from Duke Nicholas School of the Environment, DARPA TIAMAT program under award HR00112490419, and ARO under award W911NF2410405.}}

\begin{document}

\maketitle
\thispagestyle{empty}
\pagestyle{empty}

\begin{abstract}
    Heterogeneous robot teams distribute complementary capabilities across specialized agents, but their physical roles and capacities typically remain fixed throughout a mission. We present HARP, a \underline{\textbf{H}}eterogeneous \underline{\textbf{A}}erial \underline{\textbf{R}}obotic modules \underline{\textbf{P}}latform in which independently deployable aerial robots physically reconfigure to compose their capabilities for field operations. HARP comprises sensor-equipped scouts, fly-drive rover modules, and task-specific payload modules. Scouts map the environment and inform an energy-aware planner that jointly selects routes and air-ground mobility modes. Rover and payload modules fly independently across terrain that constrains ground travel, then autonomously assemble into a cooperative ground vehicle for energy-efficient payload transport. Motivated by environmental sampling in remote and difficult-to-traverse regions, we evaluate HARP through field experiments spanning sensing, planning, reconfiguration, air-ground mobility, payload transport, and task execution. We further conduct module-level deployment tests on the Greenland Ice Sheet toward future autonomous missions. HARP demonstrates how heterogeneous robot teams can adapt not only their actions, but also how their physical capabilities are composed during a mission.
\end{abstract}


\section{INTRODUCTION}
Autonomous robots are increasingly deployed for environmental monitoring \cite{rossello2021information,liu2023localization,yang2025hierarchical}, infrastructure inspection \cite{nikolic2013uav}, disaster response \cite{khan2022emerging}, and field exploration \cite{liu2025wildfusion}. However, their embodiments are typically optimized for a limited range of operating conditions. Size, weight, and power (SWaP) constraints impose fundamental tradeoffs among sensing, mobility, payload capacity, and endurance. Aerial robots can rapidly traverse challenging terrain but consume substantial energy in flight, whereas ground robots travel more efficiently and carry heavier payloads but are constrained by terrain. These tradeoffs become particularly limiting in field missions that require large-area sensing, long-distance traversal, and deployment of task-specific payloads.

Heterogeneous robot teams address part of this challenge by distributing complementary functions across specialized agents. Air-ground and legged-aerial systems combine agile flight with the greater endurance of a ground carrier \cite{lo2024experimental,depetris2022marsupial,lindqvist2022multimodality}, dockable UAV-UGV systems support coupled and independent operation \cite{deng2023selfspin,narvaez2020autonomous}, and heterogeneous teams have demonstrated long-duration autonomy in challenging field environments \cite{carlson2023multiday,oberacker2026mosaic,lindqvist2022multimodality,tranzatto2022cerberus,richter2026practical}. Yet the physical roles and capacities of individual robots generally remain fixed. Robots can share information, coordinate actions, and allocate tasks, but they rarely compose their physical capabilities. A mission requiring actuation or payload capacity beyond that of an individual robot therefore will struggle and require a larger dedicated platform.
\begin{figure}[t]
    \centering
    \includegraphics[width=0.98\columnwidth]{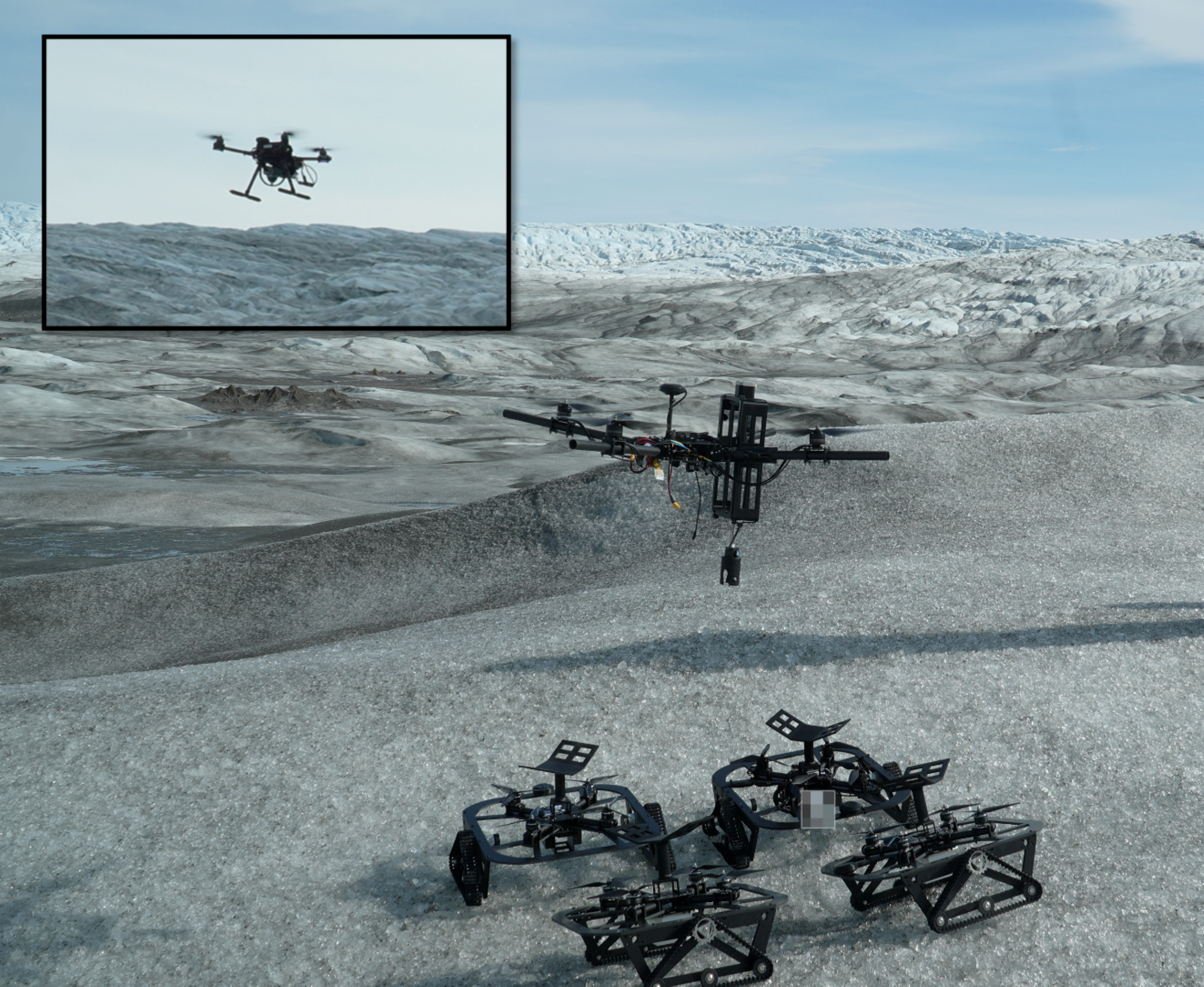}
    \vspace{-8pt}
    \caption{HARP modules undergoing unit field tests at the margin of the Greenland Ice Sheet near Kangerlussuaq Greenland. The system has three types of aerial modules: scouts for terrain mapping and path planning, rovers for ground traversal and payload transport, and payloads for mission-specific tasks.}
    \label{fig:greenland}
    \vspace{-12pt}
\end{figure}

\begin{figure*}[t]
    \centering\includegraphics[width=0.99\textwidth,height=0.4\textheight,keepaspectratio]{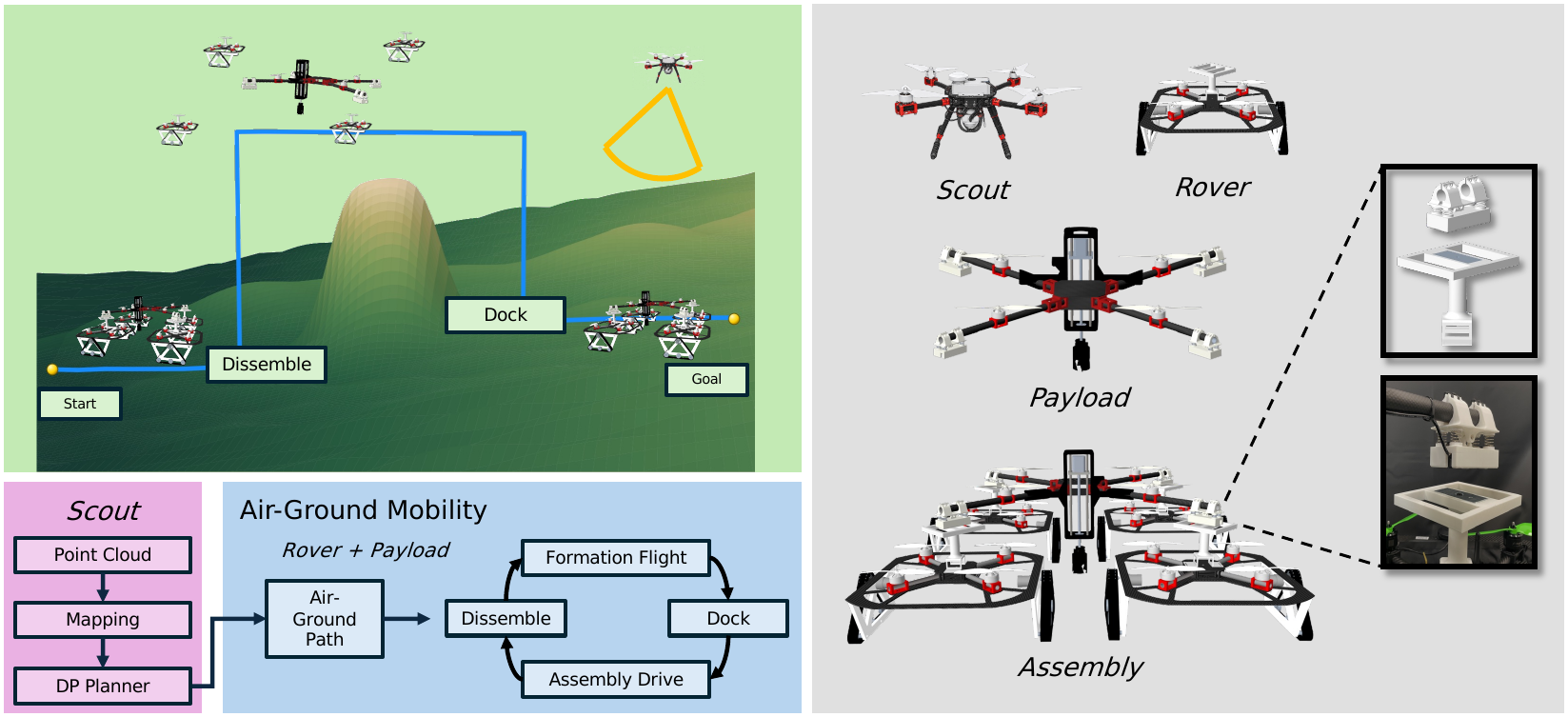}
    \vspace{-6pt}
    \caption{Overview of the HARP system architecture and modular configurations.}
    \label{fig:harp-system-overview}
    \vspace{-12pt}
\end{figure*}

Modular robots provide a complementary approach in which robots physically connect to form systems with capabilities beyond those of their constituent modules \cite{yim2000polybot,murata2002m,liang2020freebot,salemi2006superbot,romanishin2013m}. Prior work has established mechanisms, control strategies, and autonomous behaviors for reconfigurable robots, including modular aerial systems. However, physical reconfiguration has largely been studied as a capability of the robot itself instead of part of an integrated field-autonomy pipeline involving environmental perception, mobility planning, heterogeneous task roles, and mission execution. This motivates heterogeneous robots that can operate independently when advantageous and physically compose their capabilities when mission demands exceed those of individual agents.

In this work, we present \textbf{HARP}, a \textbf{H}eterogeneous \textbf{A}erial \textbf{R}obotic modules \textbf{P}latform (Fig.~\ref{fig:greenland}) for autonomous reconfigurable air-ground operations in unstructured field environments. HARP consists of three complementary aerial module types: sensor-equipped scouts, fly-drive rovers, and task-specific payload modules. All modules can deploy independently by air, while rover and payload modules can autonomously assemble into a cooperative ground vehicle. This physical reconfiguration composes multiple rovers for energy-efficient payload transport while retaining independent flight when aerial mobility is advantageous.

HARP couples this physical architecture with perception, planning, and coordinated control (Fig.~\ref{fig:harp-system-overview}). A scout maps the terrain and provides an environment representation to an energy-aware planner that jointly selects the route and air-ground mobility mode. The rover-payload team executes the resulting plan by transitioning between independent formation flight and assembled ground traversal through autonomous docking and disassembly. HARP therefore allows the team to adapt not only its trajectory and task allocation, but also how its physical capabilities are composed over the course of a mission.

We evaluate HARP through outdoor field experiments spanning terrain-aware planning, physical reconfiguration, cooperative payload transport, and integrated mission execution with a drilling payload for subsurface sampling. Motivated by remote environmental sampling, we also conduct preliminary module-level tests on the Greenland Ice Sheet to inform future field deployments. The main contributions of this work are:
{\setlength{\leftmargini}{1.5em}
\begin{itemize}
    \setlength{\topsep}{0pt}
    \setlength{\itemsep}{1pt}
    \setlength{\parskip}{0pt}
    \setlength{\parsep}{0pt}
    \item We present a heterogeneous modular robotic system integrating aerial sensing, independent flight, and reconfigurable ground traversal for cooperative transport;
    \item We develop an autonomy framework for terrain-aware air-ground planning, coordinated flight, docking, and ground motion; and
    \item We conduct experimental validation of sensing, planning, reconfiguration, payload transport, and end-to-end outdoor field operation.
\end{itemize}

\section{RELATED WORK}
\noindent\textbf{Heterogeneous Air-Ground Robotic Systems.} Heterogeneous robot teams distribute sensing and task capabilities across specialized agents while combining complementary mobility, endurance, and payload capacity for large-scale field operations \cite{orekhov2023inspiring}. Prior efforts have primarily addressed inter-robot state estimation \cite{faessler2014monocular}, communication \cite{deng2023distributed}, planning \cite{lee2012autonomous,falanga2017vision}, and coordination \cite{korsah2013comprehensive,choi2009consensus,hyun2025crew}, enabling aerial and ground robots to share observations and coordinate their actions.

At larger scales, heterogeneous teams have demonstrated cooperative mapping, exploration, inspection, and long-duration autonomy in unstructured field environments, including subterranean exploration \cite{roucek2021system,agha2022nebula,tranzatto2022cerberus}, multiday deployments \cite{carlson2023multiday}, and large-scale autonomy in extreme climates and terrain \cite{oberacker2026mosaic,richter2026practical}. These systems demonstrate the value of assigning sensing, mobility, and task roles to different robots. However, their physical embodiments and carrier-payload relationships generally remain fixed during operation. Cooperation changes how robots share information and tasks, but not how their actuation and payload capacities are physically composed. HARP complements this paradigm by allowing independently deployable aerial modules to assemble into cooperative ground vehicle when collective physical capacity is advantageous.

\noindent\textbf{Modular and Reconfigurable Robotic Systems.} Modular robots alter their configuration by connecting with other units, forming larger systems with capabilities beyond those of any individual module \cite{seo2019modular}. Early systems established this concept across a range of modular architectures \cite{yim2000polybot,murata2002m,salemi2006superbot,romanishin2013m}. More recent work has strengthened their potential for field deployment through improved deployability and autonomous, perception-driven operation \cite{liang2020freebot,daudelin2018integrated,roehr2014reconfigurable,wang2008force,neppel2025designing}.

This concept has been extended beyond ground robotics to other robotic domains. Modules assemble into larger flying structures to expand collective capabilities or improve flight performance \cite{saldana2018modquad,oung2011distributed,duffy2015lift,carlson2022armvtol}. Related multimodal robots combine aerial and terrestrial locomotion, while cooperative transport systems distribute payloads across multiple agents \cite{sihite2023morphobot,zhu2024takeoff,liu2023transportation}. These works demonstrate that physical composition can expand mobility, actuation, and payload capabilities beyond those available to an individual robot. HARP instead uses physical reconfiguration as part of a heterogeneous air-ground mission, allowing independently deployable modules to assemble when collective ground mobility and payload capacity are advantageous.

\section{METHOD}
\subsection{System Overview}
Fig.~\ref{fig:harp-system-overview} summarizes the HARP hardware architecture, autonomy stack, and information flow. A scout uses LiDAR-based simultaneous localization and mapping (LiDAR-SLAM) to map terrain locally and utilizes the Global Positioning System (GPS) receiver to transform the map into Earth-referenced GPS coordinates. A centralized dynamic-programming (DP) planner computes modality-specific cost-to-go values to jointly select a route and mobility mode, assigning assembled ground traversal or independent flight to different path segments. During aerial segments, the payload and rover modules fly in formation. At a planned transition site, the rovers establish the docking geometry and the payload performs a controlled vertical landing to assemble the team into a ground vehicle. The reverse sequence returns the modules to independent flight. In ground mode, a group-level rigid-body controller maps the desired motion of the assembled system to individual rover commands.

\subsection{Module and Mechanical Design}

HARP comprises three functionally distinct aerial modules: scout, rover, and payload. They share independent flight capability while performing different roles during a mission (Fig.~\ref{fig:harp-system-overview}). The rover and payload modules additionally share a mechanical interface that allows four rovers to assemble with one payload module into a cooperative ground vehicle.

\noindent\textbf{Scout.} The scout measures $40\times40\times25~\mathrm{cm}$ with a takeoff mass of $1.85~\mathrm{kg}$. It is powered by a 6S, 5200-mAh LiPo battery ($22.2~\mathrm{V}$ nominal) and uses BrotherHobby Avenger 2812 V3 900-KV motors, $9\times4.5$-in tri-blade propellers, and a Lumenier ELITE PRO 60A 2--6S AM32 4-in-1 ESC. Its sensing stack includes a Livox Mid-360 LiDAR, a u-blox ZED-F9P real-time kinematic GPS (RTK-GPS) receiver, and the ICM-42688-P and BMI088 inertial sensors integrated into a Holybro Pixhawk 6C Mini Model A flight controller. LiDAR processing and mapping run on a LattePanda Alpha S/864s companion computer with an Intel Core m3-8100Y processor.

\noindent\textbf{Rover.} Each rover measures $50\times50\times24~\mathrm{cm}$, weighs $3.7~\mathrm{kg}$, and incorporates the payload-support rod. Its aerial propulsion system uses a 6S, 3300-mAh LiPo battery, BrotherHobby Avenger 3115 V5 910-KV motors, $9\times4.5$-in two-blade propellers, and the same 60-A ESC as the scout. Ground locomotion is provided by two tracks, each $4.2~\mathrm{cm}$ wide with a $24.5~\mathrm{cm}$ ground-contact length. The track center-to-center spacing is $55.8~\mathrm{cm}$ and the ground clearance of $11.4~\mathrm{cm}$. Each track is driven by a 12-V Hiwonder JGB37-520R90-12 geared motor with a 90:1 reduction and rated speed of 85~rpm. The fly-drive frame is constructed from customized $4.0~\mathrm{mm}$ carbon-fiber composite plates. An Orange Pi 5 Max with a Rockchip RK3588 SoC provides onboard computation, while a Holybro Pixhawk 6C Mini Model A and RTK-GPS receiver provide flight control and positioning.

\noindent\textbf{Payload module.} The payload module measures $90\times90\times34~\mathrm{cm}$ and has a total mass of $5.1~\mathrm{kg}$. It uses a 6S, 5200-mAh LiPo battery, BrotherHobby Avenger 3115 V5 730-KV motors, $11\times4.5$-in two-blade propellers, and the same 60A ESC as above. As on the rover modules, onboard computation runs on an Orange Pi, flight control uses a Holybro Pixhawk 6C Mini Model A, and an RTK-GPS receiver provides positioning. The payload frame supports interchangeable task hardware. In this work, it carries a drilling mechanism for subsurface sampling.

\noindent\textbf{Docking and ground assembly.} The four rover modules attach one-to-one to the four arms of the payload quadrotor, forming a symmetric four-point support configuration after assembly. Distributing propulsion across four independently driven rover units improves terrain contact and provides a 4WD-like architecture compatible with traction- and slip-aware rough-terrain control \cite{kim2016kinematic}. Each rover connects to one payload arm through a 12-V power-to-lock electromagnetic interface with a seller-rated normal holding force of $60~\mathrm{kgf}$ ($\approx588~\mathrm{N}$) per lock. Passive compliance is incorporated at each connection. Springs distribute vertical load. A pitch hinge allows the rover to conform to local terrain slope. Axial clearance accommodates small relative displacements without overconstraining the assembly. Together, these features help maintain track contact on uneven terrain without additional suspension actuators. The assembled system measures 141$\times$137$\times$49~$\mathrm{cm}$ and weighs 19.9~$\mathrm{kg}$. The landing guides tolerate up to $10~\mathrm{cm}$ of lateral error before electromagnetic locking.

\subsection{Mapping and Air-Ground Path Planner}
The LiDAR cloud is rasterized into a 2.5-D map. For grid cell $\mathcal{C}_{ij}$, its elevation samples are
\begin{equation}
    \mathcal{P}_{ij}=\{z_n:(x_n,y_n,z_n)\in\mathcal{P},
    (x_n,y_n)\in\mathcal{C}_{ij}\}
    \label{eq:point-cell-set}
\end{equation}
Ground elevation $h_{ij}$ is the smoothed 10th-percentile return, and roughness is $r_{ij}=Q_{0.75}(\mathcal{P}_{ij})-Q_{0.25}(\mathcal{P}_{ij})$. With obstacle cells $\mathcal{O}$ and unsafe landing cells $\mathcal{L}$, these define
\begin{equation}
    \mathcal{M}=\{h_{ij},r_{ij},\mathcal{O},\mathcal{L}\},
    \label{eq:planner-map}
\end{equation}
where $\mathcal{O}$ thresholds surface height and slope; $\mathcal{L}$ additionally thresholds roughness. Unobserved cells permit neither ground travel nor landing.

Route and mobility mode are jointly planned on an eight-connected lattice with state $q=(i,j,m)$, where $m\in\{\mathrm{g},\mathrm{a}\}$ denotes assembled ground travel or disassembled flight. Ground edges are feasible only when the swept system footprint remains collision-free and satisfies attitude constraints. Aerial edges maintain prescribed terrain clearance, while mode-transition states require a collision-free assembled footprint outside $\mathcal{L}$. Heading and bookkeeping states enforce mode-use and minimum-flight-leg constraints. Edge costs approximate energy during ground motion, flight, and reconfiguration. Reverse Dijkstra search from the ground-mode targets produces the cost-to-go map and minimum-energy route. The full cost model is given in Appendix~\ref{app:planner-cost}.

\subsection{Formation Flight}
After the DP planner produces a modality-labeled path, the rover-payload team executes it segment by segment. For aerial path segments, the payload acts as the formation leader while the $N=4$ rover modules fly independently at prescribed offsets. Maintaining this formation places the rovers near their required docking geometry at the subsequent transition site.

Let $d=3$ denote the Cartesian dimension. The payload position is $\mathbf{p}_0\in\mathbb{R}^{d}$, and $\mathbf{p}_i\in\mathbb{R}^{d}$ is the position of rover $i\in\{1,\ldots,N\}$ in a shared inertial frame. The constant vector $\mathbf{r}_i\in\mathbb{R}^{d}$ specifies rover $i$'s desired offset from the payload. Information exchange is represented by the leader-follower graph
\begin{align}
    \mathcal{G} &= (\mathcal{V},\mathcal{E}), \notag\\
    \mathcal{V} &= \{0,1,\ldots,N\},
    \qquad N=4,
    \label{eq:formation-graph}
\end{align}
where node $0$ is the payload and node $i$ is rover $i$. An edge $(j,i)\in\mathcal{E}$ indicates that rover $i$ receives the state of node $j$. The follower weight satisfies $a_{ij}>0$ when the corresponding edge exists and $a_{ij}=0$ otherwise. Similarly, $b_i=1$ when rover $i$ directly receives the payload state and $b_i=0$ otherwise. The graph contains a directed spanning tree rooted at the payload.

The position consensus error for rover $i$ is
\begin{align}
    \mathbf{e}_i ={}& \sum_{j=1}^{N} a_{ij}
    \left[(\mathbf{p}_i-\mathbf{r}_i)
    -(\mathbf{p}_j-\mathbf{r}_j)\right] \notag\\
    &+b_i(\mathbf{p}_i-\mathbf{r}_i-\mathbf{p}_0)
    \in\mathbb{R}^{d}.
    \label{eq:formation-position-error}
\end{align}
Under the leader-rooted graph, $\mathbf{e}_i=\mathbf{0}_{d}$ for all followers corresponds to the desired formation $\mathbf{p}_i=\mathbf{p}_0+\mathbf{r}_i$. The stacked derivation, velocity error, Proportional-Derivative (PD) control law, and gain conditions are provided in Appendix~\ref{app:formation-control}.

\subsection{Vertical Docking and Reconfiguration}
At a ground-transition site, the four rovers first establish the docking geometry defined by the payload arms. Before each payload landing, rover~1 acts as the local formation leader and aligns its yaw with the planned ground-path heading. The remaining three rovers independently regulate their assigned offsets using the consensus error in Eq.~\eqref{eq:formation-position-error}. During this phase, the leader term is supplied by rover~1, and the payload is excluded from the formation graph.

Once the rover formation converges, its geometric center in the horizontal plane becomes the payload's landing setpoint. The payload tracks this center and performs a controlled vertical descent onto the landing supports. Upon touchdown, the four electromagnetic locks engage and mechanically secure the payload to the rover assembly and complete the transition from distributed flight to assembled ground operation. Disassembly follows the reverse sequence. The locks release, the payload takes off vertically, and the rover-payload team returns to independent aerial operation.

\subsection{Coordinated Ground Motion}
After docking, we model the payload and the $N=4$ rover modules as a single planar rigid body. Let $\mathcal{F}_{0}$ be a payload-fixed frame centered at the payload, retaining the payload index $0$ from the formation controller. In this frame, the desired payload twist is
\begin{align}
    \boldsymbol{\xi}_{0}^{d}
    &= \begin{bmatrix}
        (\mathbf{v}_{0,\mathrm{g}}^{d})^{\top} & \omega_{0}^{d}
    \end{bmatrix}^{\top}
    \in\mathbb{R}^{3}, \notag\\
    \mathbf{v}_{0,\mathrm{g}}^{d}
    &= \begin{bmatrix}v_x^{d} & v_y^{d}\end{bmatrix}^{\top}
    \in\mathbb{R}^{2},
    \label{eq:payload-planar-twist}
\end{align}
where $\omega_{0}^{d}\in\mathbb{R}$ is the desired yaw rate. For rover $i\in\{1,\ldots,N\}$, let $\boldsymbol{\rho}_i=[x_i,y_i]^{\top}\in\mathbb{R}^{2}$ denote its fixed attachment vector from the payload center, expressed in $\mathcal{F}_{0}$. These vectors encode the four-rover geometry shown in Fig.~\ref{fig:harp-system-overview}. The planar rigid-body velocity relation gives the desired rover velocity as
\begin{equation}
\begin{aligned}
    \mathbf{v}_{i,\mathrm{g}}^{d}
    &= \mathbf{v}_{0,\mathrm{g}}^{d}
    +\omega_{0}^{d}\mathbf{J}\boldsymbol{\rho}_i \in\mathbb{R}^{2}.
    %  \\
    % &= \begin{bmatrix}
    %     v_x^{d}-\omega_{0}^{d}y_i\\
    %     v_y^{d}+\omega_{0}^{d}x_i
    % \end{bmatrix}
\end{aligned}
    \label{eq:rigid-body-rover-velocity}
\end{equation}
Here, $\mathbf{J}=\left[\begin{smallmatrix}0&-1\\1&0\end{smallmatrix}\right]\in\mathbb{R}^{2\times2}$ is the planar skew-symmetric matrix, and $\omega_{0}^{d}\mathbf{J}\boldsymbol{\rho}_i$ is the tangential velocity induced by rotation about the payload center. All modules share the desired yaw rate $\omega_{0}^{d}$. The allocation assumes fixed attachment geometry. Passive connector compliance and track-terrain slip are not explicitly modeled.

% Queue Figs. 4--6 in numerical order before the Results section.
\begin{figure*}[!t]
    \centering
    \includegraphics[width=\textwidth]{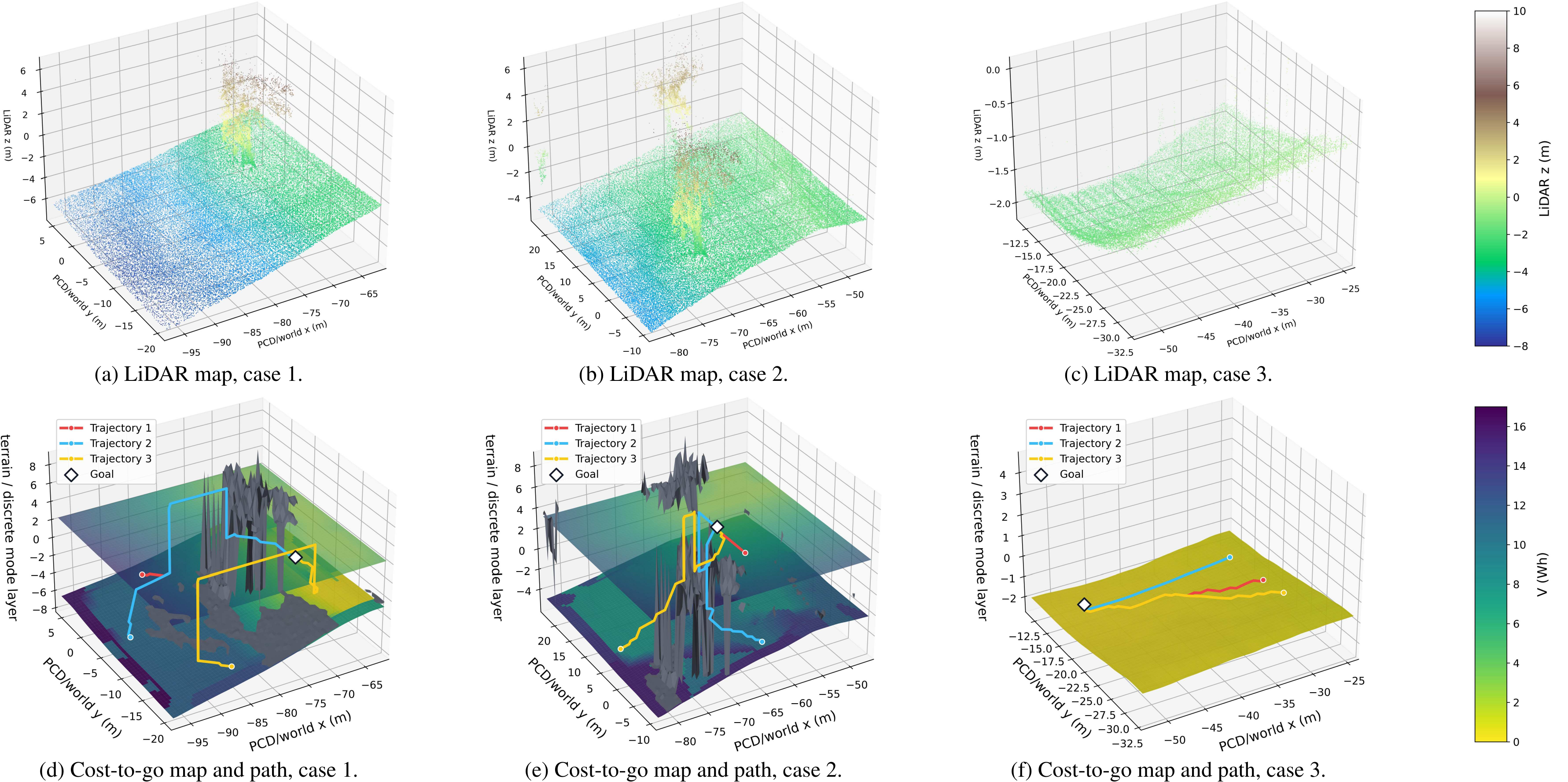}
    \caption{Terrain mapping and joint air-ground planning across three representative environments. Top: scout-generated LiDAR maps colored by elevation. Bottom: cost-to-go $V$ (yellow low, blue--purple high) over the ground and aerial mobility layers with blocked states shown in gray. Red, cyan, and yellow denote routes and initial positions 1--3, and the white diamond is the shared ground goal. Vertical segments indicate mode transitions.}
    \label{fig:planned-air-ground-path}
\end{figure*}

\begin{figure}[!t]
    \centering
    \subfloat[Mean pre-task payload energy across three initial conditions.\label{fig:power_consumption}]{%
        \includegraphics[width=0.95\columnwidth]{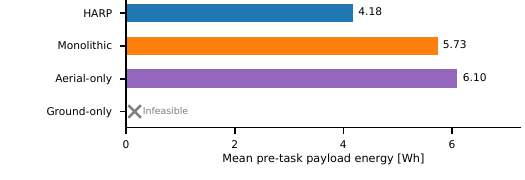}%
    }\\[0.5em]
    \subfloat[Measured payload weight carried by one rover and four rovers.\label{fig:payload_comparison}]{%
        \includegraphics[width=0.95\columnwidth]{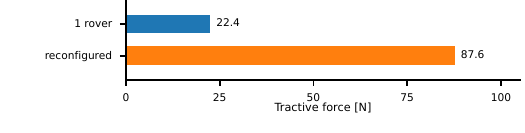}
    }
    \caption{Physical benefits of reconfiguration. (a) Model-estimated energy drawn from the payload or integrated task-bearing platform before task execution, averaged across three initial conditions. The ground-only baseline has no feasible route under the terrain and collision constraints. (b) Measured tractive force by a single rover and by the four-rover assembled configuration.}
    \label{fig:reconfigurability-comparison}
    \vspace{-8pt}
\end{figure}

\begin{figure}[!t]
    \centering
    \includegraphics[width=0.7\columnwidth]{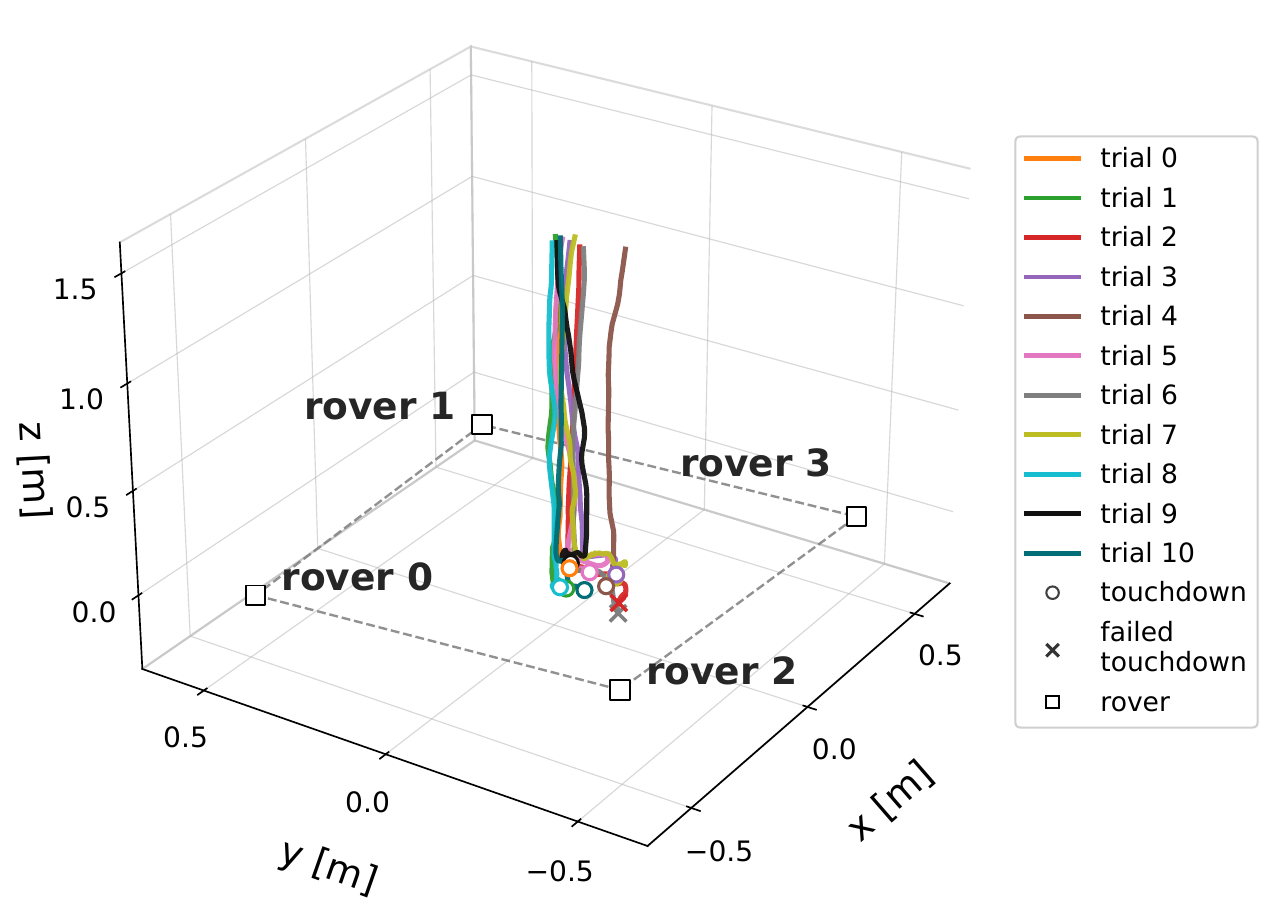}
    \vspace{-8pt}
    \caption{Autonomous vertical-docking performance over 11 repeated trials. Curves show payload trajectories during terminal descent toward the center of the four-rover formation. Circles denote successful capture and locking, while crosses denote unsuccessful trials.}
    \label{fig:docking-trajectories}
    \vspace{-8pt}
\end{figure}

\begin{figure*}[!t]
    \centering
    \begin{minipage}{0.96\textwidth}
    \centering
    \subfloat[\label{fig:field-map-path}]{%
        \parbox[b][3.5cm][c]{0.55\linewidth}{%
            \centering
            \includegraphics[
                width=\linewidth,
                height=3.5cm,
                keepaspectratio
            ]{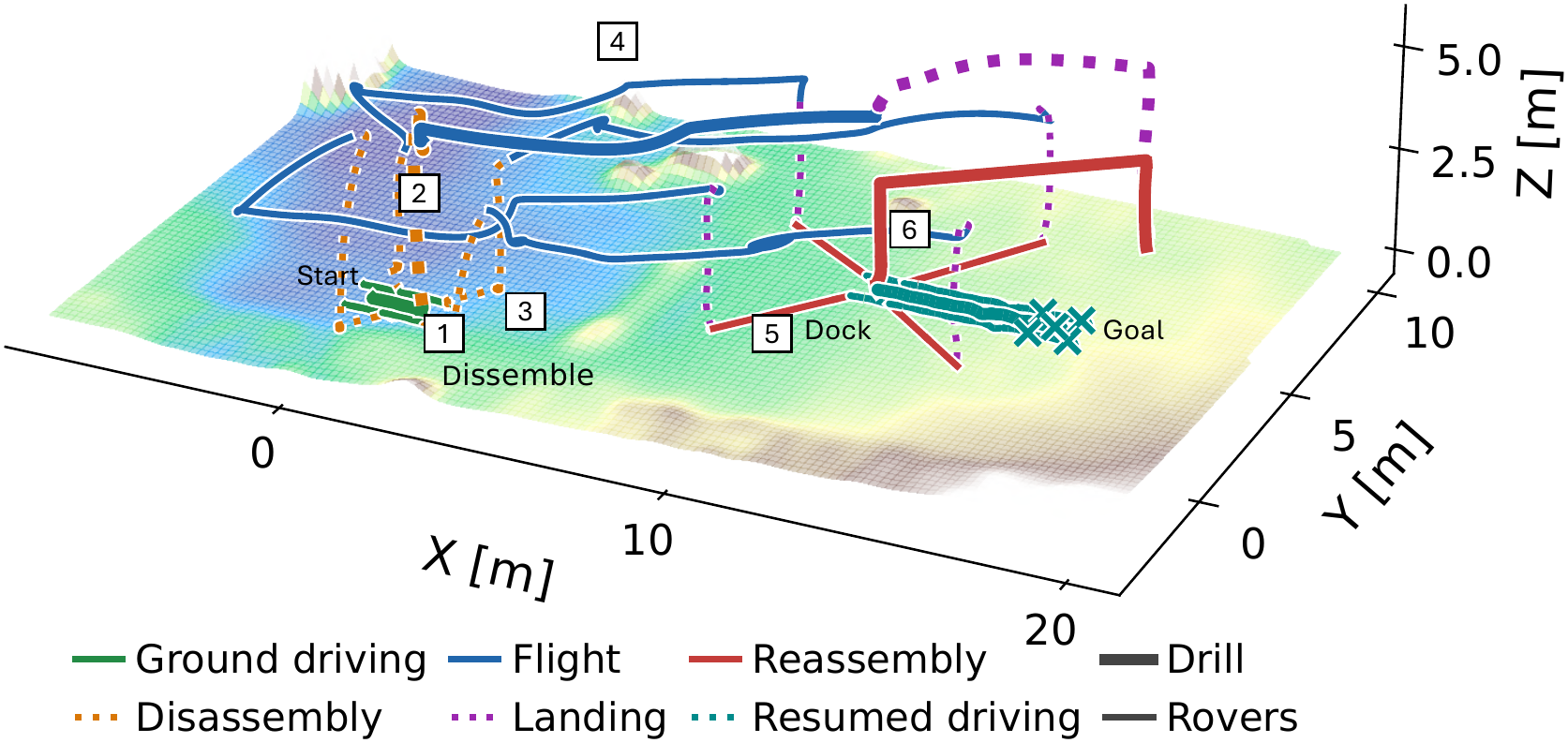}%
        }%
    }
    \hfill
    \subfloat[\label{fig:field-sample}]{%
        \parbox[b][3.5cm][c]{0.45\linewidth}{%
            \centering
            \includegraphics[
                width=\linewidth,
                height=3.5cm,
                keepaspectratio
            ]{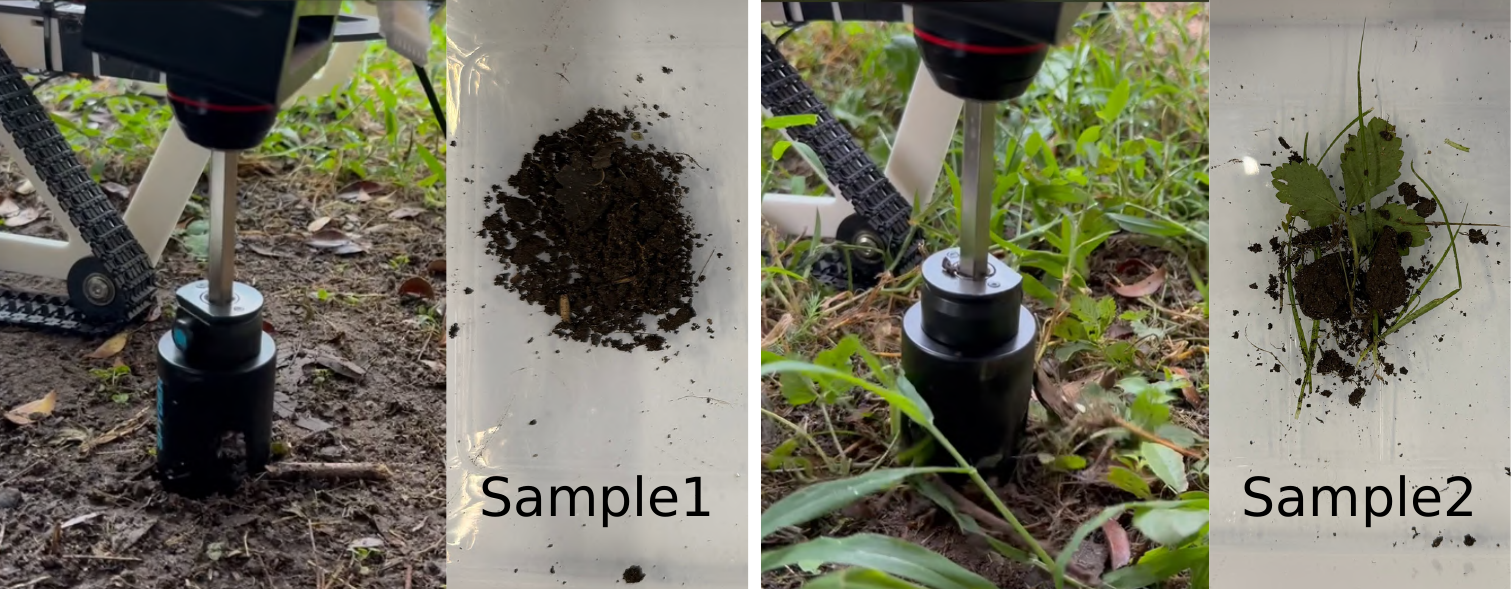}%
        }%
    }
    \end{minipage}\\[0.5em]
    \subfloat[\label{fig:field-sequence}]{%
        \includegraphics[
            width=0.96\textwidth,
            clip
        ]{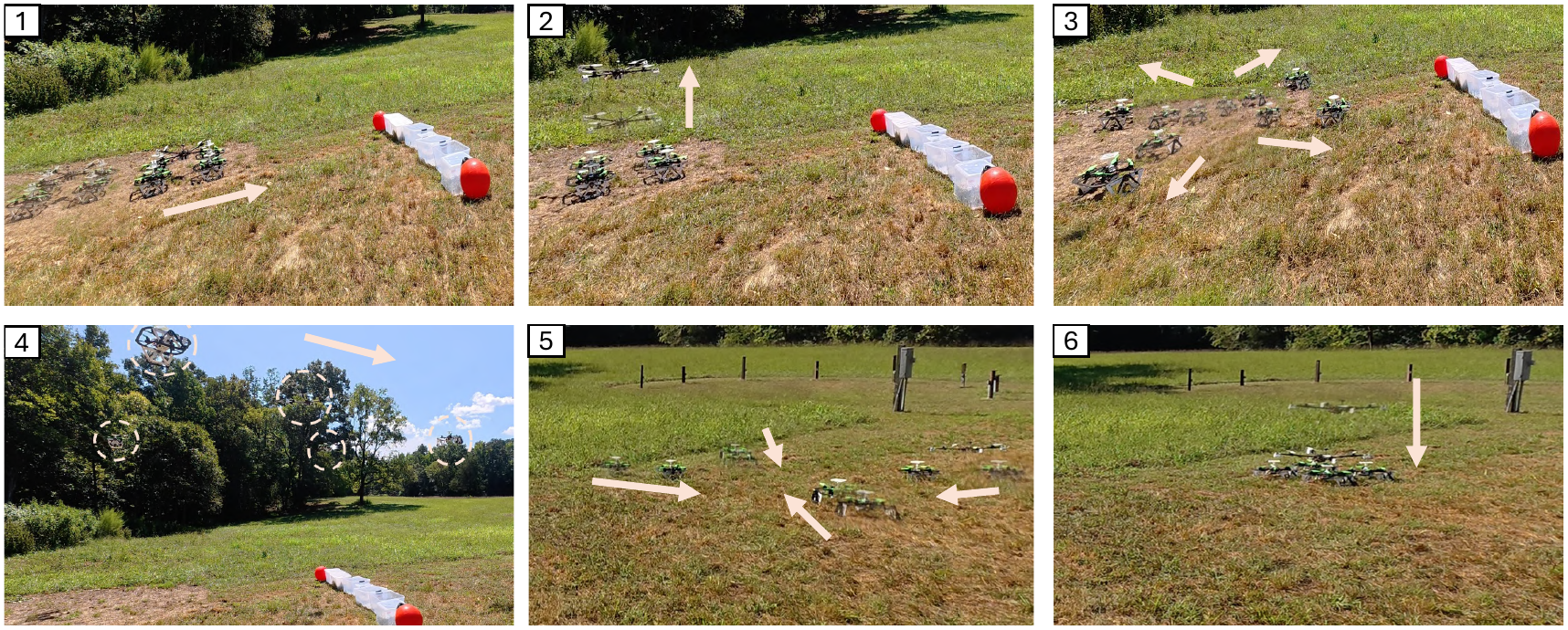}%
    }
    \caption{End-to-end outdoor field demonstration. (a) Executed trajectory over the reconstructed terrain map, with assembled ground and independent aerial segments. (b) Subsurface sample recovered by the drilling payload. (c) Representative sequence showing ground traversal, disassembly, formation flight, vertical docking, and reassembly.}
    % at the Duke Teaching Observatory
    \label{fig:field-trajectory-map}
    \vspace{-14pt}
\end{figure*}

\section{Experiments}
\subsection{Terrain Mapping and Air--Ground Planning}
We first evaluated whether the planner selected mobility modes according to terrain reconstructed from real field data by the scout. The scout used FAST-LIO~\cite{xu2021fast} for LiDAR--inertial mapping, and Fig.~\ref{fig:planned-air-ground-path} shows three representative terrain maps with the corresponding hybrid cost-to-go functions and planned routes. In the lower row, the lower and upper surfaces represent assembled ground traversal and independent aerial motion, respectively. Color indicates the cost-to-go, while gray regions denote infeasible states.

The resulting routes adapted their mobility modes to the mapped terrain. In case~2, continuous ground access allowed all three routes to remain in the assembled ground mode. In case~1 and~3, obstacles and elevation discontinuities made portions of the ground layer infeasible, causing the planner to introduce aerial segments before returning to ground operation near the goal. Thus, instead of prescribing a fixed sequence of locomotion modes, the planner selected where physical reconfiguration was beneficial based on the mapped terrain from the scout and estimated energy cost.

\subsection{Benefits of Physical Reconfiguration}
\noindent\textbf{Energy Consumption}. We evaluated the energy benefit of physical reconfiguration through numerical simulation via the air-ground planner. HARP was compared against three non-reconfigurable baselines: (1) monolithic air--ground, (2) aerial-only, and (3) ground-only systems. Each platform's edge cost is defined by its power consumption, modeled using physical parameters including mass, battery characteristics, and motor specifications. For all configurations, routes were independently planned on the Case~1 terrain in Fig.~\ref{fig:planned-air-ground-path} (bottom-left panel) using the same three initial positions and target.

Across the three initial conditions, HARP consumed a mean of $4.18~\mathrm{Wh}$ from the payload module before task execution, compared with $5.73~\mathrm{Wh}$ for the monolithic platform and $6.10~\mathrm{Wh}$ for the aerial-only baseline (Fig.~\ref{fig:power_consumption}). These values correspond to reductions of $27.1\%$ and $31.5\%$, respectively. The ground-only carrier could not reach the target under the terrain and collision constraints. Within this model, HARP reduced pre-task energy expenditure relative to the two feasible non-reconfigurable baselines while retaining reachability unavailable to the ground-only carrier.

\noindent\textbf{Tractive Force}. We measured tractive force on a controlled mat surface with the motors commanded at full actuation. A single rover generated 22.4~$\mathrm{N}$ and the docked four-rover configuration achieved 87.6~$\mathrm{N}$, a 3.91$\times$ increase (Fig.~\ref{fig:payload_comparison}), demonstrating the benefit of assembled actuation.

\subsection{Autonomous Docking Performance}
Reliable docking is required to transition repeatedly between independent flight and assembled ground operation. We therefore evaluated repeated landing trials, reporting touchdown accuracy and capture success rate. Fig.~\ref{fig:docking-trajectories} overlays the three-dimensional payload trajectories from approach through touchdown, with the center of the rover formation defined as the target. Landing accuracy was measured by the horizontal distance between each touchdown position and this target. Across 11 trials, the mean touchdown error was $0.11\pm0.04$~$\mathrm{m}$, with a maximum error of 0.17~$\mathrm{m}$.

A trial was considered successful when the horizontal touchdown error relative to the rover-formation center was less than $0.15\,\mathrm{m}$ and the magnetic locking mechanism successfully engaged. Nine of the 11 trials resulted in successful docking, corresponding to a success rate of $81.8\%$. Trajectory dispersion characterized landing repeatability. The success rate reflected the combined effects of state estimation, formation control, landing accuracy, and mechanical capture tolerance. These metrics quantify docking reliability and identify failure modes that can guide improvements to the landing controller and support geometry.
% \vspace{-20pt}

\subsection{End-to-End Outdoor Field Demonstration}
\textbf{End-to-End Field Demonstration}. We performed an end-to-end field test in which the system executed ground traversal, disassembly, formation flight, vertical docking, reassembly, and resumed ground travel.
State estimates from all aerial modules were synchronized in a shared inertial frame. High-level states and commands over a local area network (LAN) using Robot Operating System~2 (ROS~2). Companion computers interfaced with PX4 through MAVLink. The system achieved an autonomy ratio (AR) \cite{oberacker2026mosaic} of 90.75\%, with human intervention required only for final formation adjustment before docking.

% at the Duke Teaching Observatory
We evaluated the integrated system on an outdoor test course measuring 8$\times$18~$\mathrm{m}$ and containing artificial obstacles that constrained ground travel and motivated transitions between locomotion modes. The complete mission covered a horizontal length of 26.12~$\mathrm{m}$. The assembled system traversed an initial ground segment, disassembled into independent aerial modules, flew across the obstacle region to a new transition site, landed, reassembled, and resumed ground traversal. Of the total route, 3.42~$\mathrm{m}$ was completed in ground mode and 22.7~$\mathrm{m}$ in aerial mode. The mission included 2 mobility/reconfiguration transitions, located at $(1.31, 0.048)$, $(10.79, 4.74)$~$\mathrm{m}$ within the local inertial frame. 

This end-to-end sequence demonstrates coordinated reconfiguration and multimodal mobility outside a controlled indoor setting. Fig.~\ref{fig:field-map-path} shows the system trajectory over the reconstructed 2.5-D terrain map, including the artificial obstacles and the ground and aerial segments of the route. Fig.~\ref{fig:field-sequence} presents consecutive frames of the demonstration, showing the corresponding ground traversal, disassembly, flight, landing, reassembly, and resumed ground traversal.

\textbf{Environmental Sampling}. Following the mobility demonstration, the payload drill successfully collected soil from humid ground, as shown in Fig.~\ref{fig:field-sample}. The current design could not reliably extract a complete specimen or penetrate harder soil, but a more capable drill is under development for future field testing and, ultimately, firn sampling.

% \vspace{-10pt}
\subsection{Module-Level Field Tests on the Greenland Ice Sheet}
In addition to the integrated trials shown above, we transported individual modules to Kangerlussuaq, Greenland, to assess deployment logistics and module-level operation in the target environment, as shown in Fig. \ref{fig:greenland}. All modules completed individual flight tests, and the rover modules additionally traversed the local terrain. These tests provided an initial assessment of the individual modules outside the primary experimental setting and informed the design requirements for future full-scale field deployment.
% \vspace{-10pt}
% We are now extending the system toward large-scale autonomous deployment. This effort targets more reliable reconfiguration and coordination under wind, uneven terrain, and intermittent communication. Near-term improvements include compliant capture guides and a mechanically retained, weather-resistant connector, relative-pose feedback during terminal landing, and traffic-aware task allocation and conflict-free path scheduling for larger teams. Redundant local networking with a telemetry fallback can further maintain coordination when communication links degrade.

\section{Conclusions, Limitations, and Future Work}
We presented HARP, a heterogeneous modular robotic platform in which independently deployable aerial robots physically reconfigure into a cooperative ground vehicle. By integrating terrain sensing and energy-aware air-ground planning with formation flight, autonomous docking, and coordinated ground motion, HARP enables a robot team to adapt not only its actions, but also how its physical capabilities are composed during a mission. Outdoor experiments demonstrate terrain-dependent mobility selection, the energy and payload benefits of reconfiguration, autonomous physical assembly, and end-to-end air-ground operation with a task-specific sampling payload.

The current system remains an initial realization. Autonomous docking succeeded in 9 of 11 trials, making reconfiguration reliability a primary limitation for repeated operation, especially under harsh environmental conditions. The drilling experiment demonstrates integrated task execution in soft soil. Sampling on harder surfaces will require further evaluation of drilling stability.

Future work will focus on improving reconfiguration robustness through relative-pose feedback, compliant capture mechanisms, and mechanically retained weather-resistant connectors. Evaluating the system over longer-duration missions under extreme wind, uneven terrain, and intermittent communication is a valuable direction. Broadly, HARP provides a foundation for heterogeneous robot teams that can reorganize their physical capabilities as mission demands evolve.
% \vspace{-10pt}
\appendix[Supplementary Method Details]
\subsection{Air--Ground Planning Cost Model}
\label{app:planner-cost}
Edge costs are expressed in Wh. With payload mass $m_p$ shared by four rovers, each rover carries $m_\ell=m_r+m_p/4$. For path length $d\ell$, roughness $r$, grade $\alpha$, and speed $v$,
\begin{align}
    C_{\mathrm{rr}}&=C_0+k_r r, \notag\\
    F_{\mathrm{tr}}&=\max\{0,m_\ell g(C_{\mathrm{rr}}\cos\alpha+\sin\alpha)\}, \notag\\
    dE_{\mathrm{g}}&=\frac{1}{3600}
    \left(P_{\mathrm{e}}+\frac{F_{\mathrm{tr}}v}{\eta}\right)\frac{d\ell}{v}.
    \label{eq:ground-energy}
\end{align}
The assembled ground cost is $4\int dE_{\mathrm{g}}$. Flight power for phase $k\in\{\mathrm{to},\mathrm{cr},\mathrm{ld}\}$ is
\begin{equation}
    P_k(m)=P_{k,\mathrm{ref}}
    \left[f_0+(1-f_0)\left(\frac{m}{m_{\mathrm{ref}}}\right)^{3/2}\right].
    \label{eq:flight-power}
\end{equation}
An aerial edge of length $d$ therefore costs
$[4P_{\mathrm{cr}}(m_r)+P_{\mathrm{cr}}(m_p)]d/(3600v_{\mathrm{a}})$,
with transition energies added for takeoff, landing, docking, and undocking.

On the hybrid graph $(\mathcal{Q},\mathcal{E})$, the cost-to-go satisfies
\begin{align}
    V(q)&=0,\quad q\in\mathcal{Q}_T, \notag\\
    V(q)&=\min_{(q,q')\in\mathcal{E}}
    \left[c(q,q')+V(q')\right].
    \label{eq:dp-bellman}
\end{align}
Reverse Dijkstra search from $\mathcal{Q}_T$ computes $V(q)$, and each route follows the minimizing successor.

\subsection{Formation-Control Derivation}
\label{app:formation-control}
Let $\mathbf{A}=[a_{ij}]$ denote the follower adjacency matrix,
$\mathbf{D}=\operatorname{diag}(\sum_j a_{ij})$, and
$\mathbf{B}=\operatorname{diag}(b_i)$ the leader-pinning matrix. Define
$\mathbf{H}=\mathbf{D}-\mathbf{A}+\mathbf{B}$, which is nonsingular for a
leader-rooted graph.

Let $\mathbf{p}_{\mathrm{F}}=\operatorname{col}(\mathbf{p}_i)$,
$\mathbf{v}_{\mathrm{F}}=\operatorname{col}(\mathbf{v}_i)$, and
$\mathbf{r}=\operatorname{col}(\mathbf{r}_i)$. The stacked formation errors are
\begin{align}
    \mathbf{e}
    &= (\mathbf{H}\otimes\mathbf{I}_d)
    (\mathbf{p}_{\mathrm{F}}-\mathbf{r}
    -\mathbf{1}_N\otimes\mathbf{p}_0), \notag\\
    \dot{\mathbf{e}}
    &= (\mathbf{H}\otimes\mathbf{I}_d)
    (\mathbf{v}_{\mathrm{F}}
    -\mathbf{1}_N\otimes\mathbf{v}_0).
    \label{eq:formation-error}
\end{align}
For constant offsets $\mathbf{r}_i$, each follower applies
\begin{equation}
    \mathbf{u}_i
    =\mathbf{a}_0
    -\mathbf{K}_{\mathrm{p}}\mathbf{e}_i
    -\mathbf{K}_{\mathrm{d}}\dot{\mathbf{e}}_i,
    \qquad i=1,\ldots,N,
    \label{eq:formation-pd}
\end{equation}
where $\mathbf{K}_{\mathrm{p}},\mathbf{K}_{\mathrm{d}}\succ\mathbf{0}$.
\vspace{-0.1cm}
\section*{ACKNOWLEDGMENT}
The authors thank Maya Maciel-Seidman and Jasper Heuer for their assistance during field operations on the Greenland Ice Sheet, and Gavin Yang and Alexa Kim for their contributions to the early exploration of this project.

% Include every entry from reference.bib.
% \nocite{*}
\bibliographystyle{ieeetr}
\bibliography{reference}

@article{roehr2014reconfigurable,
  title={Reconfigurable integrated multirobot exploration system (RIMRES): Heterogeneous modular reconfigurable robots for space exploration},
  author={Roehr, Thomas M and Cordes, Florian and Kirchner, Frank},
  journal={Journal of Field Robotics},
  volume={31},
  number={1},
  pages={3--34},
  year={2014},
  publisher={Wiley Online Library}
}

@article{wang2008force,
  title={Force cooperation in a reconfigurable field multirobot system},
  author={Wang, Wei and Zhang, Houxiang and Zong, Guanghua and Zhang, Jianwei},
  journal={Journal of Field Robotics},
  volume={25},
  number={11-12},
  pages={922--938},
  year={2008},
  publisher={Wiley Online Library}
}

@inproceedings{neppel2025designing,
  title={Designing for Distributed Heterogeneous Modularity: On Software Architecture and Deployment of the MoonBots},
  author={Neppel, Elian and Karimov, Shamistan and Mishra, Ashutosh and Diaz, Gustavo H and Gozbasi, Hazal and Santra, Shreya and Uno, Kentaro and Yoshida, Kazuya},
  booktitle={2025 International Conference on Space Robotics (iSpaRo)},
  pages={79--84},
  year={2025},
  organization={IEEE}
}

@inproceedings{kim2016kinematic,
  title={A kinematic-based rough terrain control for traction and energy saving of an exploration rover},
  author={Kim, Jayoung and Lee, Jihong},
  booktitle={2016 IEEE/RSJ International Conference on Intelligent Robots and Systems (IROS)},
  pages={3595--3600},
  year={2016},
  organization={IEEE},
  doi={10.1109/IROS.2016.7759529}
}

@article{hyun2025crew,
  title={Crew-wildfire: Benchmarking agentic multi-agent collaborations at scale},
  author={Hyun, Jonathan and Waytowich, Nicholas R and Chen, Boyuan},
  journal={arXiv preprint arXiv:2507.05178},
  year={2025}
}

@article{lo2024experimental,
  title={Experimental nonrobocentric dynamic landing of quadrotor UAVs with on-ground sensor suite},
  author={Lo, Li-Yu and Li, Boyang and Wen, Chih-Yung and Chang, Ching-Wei},
  journal={IEEE Transactions on Instrumentation and Measurement},
  volume={73},
  pages={1--13},
  year={2024},
  publisher={IEEE}
}

@article{yang2025hierarchical,
  title={Hierarchical 3D Scene Graph based Semantic-Metric SLAM for Plant Inspection and Fruit Counting in Intelligent Hydroponics System},
  author={Yang, Wenyu and Liu, Kang and Tan, Zheng and Lo, Li-Yu and Wang, Yinglun and Wong, Ka-Hing and Wen, Chih-Yung},
  journal={IEEE Internet of Things Journal},
  year={2025},
  publisher={IEEE}
}

@article{rossello2021information,
  title={Information-driven path planning for UAV with limited autonomy in large-scale field monitoring},
  author={Rossello, Nicolas Bono and Carpio, Renzo Fabrizio and Gasparri, Andrea and Garone, Emanuele},
  journal={IEEE Transactions on Automation Science and Engineering},
  volume={19},
  number={3},
  pages={2450--2460},
  year={2021},
  publisher={IEEE}
}

@inproceedings{liu2025wildfusion,
  title={WildFusion: Multimodal Implicit 3D Reconstructions in the Wild},
  author={Liu, Yanbaihui and Chen, Boyuan},
  booktitle={2025 IEEE International Conference on Robotics and Automation (ICRA)},
  pages={8603--8603},
  year={2025},
  organization={IEEE}
}

@inproceedings{nikolic2013uav,
  title={A UAV system for inspection of industrial facilities},
  author={Nikolic, Janosch and Burri, Michael and Rehder, Joern and Leutenegger, Stefan and Huerzeler, Christoph and Siegwart, Roland},
  booktitle={2013 IEEE aerospace conference},
  pages={1--8},
  year={2013},
  organization={IEEE}
}

@article{khan2022emerging,
  title={Emerging UAV technology for disaster detection, mitigation, response, and preparedness},
  author={Khan, Amina and Gupta, Sumeet and Gupta, Sachin Kumar},
  journal={Journal of Field Robotics},
  volume={39},
  number={6},
  pages={905--955},
  year={2022},
  publisher={Wiley Online Library}
}

@inproceedings{yim2000polybot,
  title={PolyBot: a modular reconfigurable robot},
  author={Yim, Mark and Duff, David G and Roufas, Kimon D},
  booktitle={Proceedings 2000 ICRA. millennium conference. IEEE international conference on robotics and automation. Symposia proceedings (Cat. No. 00CH37065)},
  volume={1},
  pages={514--520},
  year={2000},
  organization={IEEE}
}

@article{agha2022nebula,
  title={NeBula: TEAM CoSTAR's robotic autonomy solution that won phase II of DARPA subterranean challenge},
  author={Agha, Ali and Otsu, Kyohei and Morrell, Benjamin and Fan, David D and Thakker, Rohan and Santamaria-Navarro, Angel and Kim, Sung-Kyun and Bouman, Amanda and Lei, Xianmei and Edlund, Jeffrey and others},
  journal={Field robotics},
  volume={2},
  pages={1432--1506},
  year={2022},
  publisher={FRPS}
}

@article{seo2019modular,
  title={Modular reconfigurable robotics},
  author={Seo, Jungwon and Paik, Jamie and Yim, Mark},
  journal={Annual Review of Control, Robotics, and Autonomous Systems},
  volume={2},
  number={1},
  pages={63--88},
  year={2019},
  publisher={Annual Reviews}
}

@inproceedings{romanishin2013m,
  title={M-blocks: Momentum-driven, magnetic modular robots},
  author={Romanishin, John W and Gilpin, Kyle and Rus, Daniela},
  booktitle={2013 IEEE/RSJ international conference on intelligent robots and systems},
  pages={4288--4295},
  year={2013},
  organization={IEEE}
}

@article{murata2002m,
  title={M-TRAN: Self-reconfigurable modular robotic system},
  author={Murata, Satoshi and Yoshida, Eiichi and Kamimura, Akiya and Kurokawa, Haruhisa and Tomita, Kohji and Kokaji, Shigeru},
  journal={IEEE/ASME transactions on mechatronics},
  volume={7},
  number={4},
  pages={431--441},
  year={2002},
  publisher={IEEE}
}

@inproceedings{salemi2006superbot,
  title={SUPERBOT: A deployable, multi-functional, and modular self-reconfigurable robotic system},
  author={Salemi, Behnam and Moll, Mark and Shen, Wei-Min},
  booktitle={2006 IEEE/RSJ International Conference on Intelligent Robots and Systems},
  pages={3636--3641},
  year={2006},
  organization={IEEE}
}

@article{choi2009consensus,
  title={Consensus-based decentralized auctions for robust task allocation},
  author={Choi, Han-Lim and Brunet, Luc and How, Jonathan P},
  journal={IEEE transactions on robotics},
  volume={25},
  number={4},
  pages={912--926},
  year={2009},
  publisher={IEEE}
}

@article{korsah2013comprehensive,
  title={A comprehensive taxonomy for multi-robot task allocation},
  author={Korsah, G Ayorkor and Stentz, Anthony and Dias, M Bernardine},
  journal={The International journal of robotics research},
  volume={32},
  number={12},
  pages={1495--1512},
  year={2013},
  publisher={SAGE Publications Sage UK: London, England}
}

@article{roucek2021system,
  title={System for multi-robotic exploration of underground environments ctu-cras-norlab in the darpa subterranean challenge},
  author={Roucek, Tom{\'a}{\v{s}} and Pecka, Martin and C{\i}zek, Petr and Petr{\i}cek, Tom{\'a}{\v{s}} and Bayer, Jan and {\v{S}}alansky, V and Azayev, Teymur and Hert, Daniel and Petrl{\i}k, Matej and B{\'a}ca, Tom{\'a}s and others},
  journal={arXiv preprint arXiv:2110.05911},
  year={2021}
}

@inproceedings{lee2012autonomous,
  title={Autonomous landing of a VTOL UAV on a moving platform using image-based visual servoing},
  author={Lee, Daewon and Ryan, Tyler and Kim, H Jin},
  booktitle={2012 IEEE international conference on robotics and automation},
  pages={971--976},
  year={2012},
  organization={IEEE}
}

@article{deng2023distributed,
  title={Distributed cooperative optimization for nonlinear heterogeneous MASs under intermittent communication},
  author={Deng, Chao and Xu, Lei and Yang, Tao and Yue, Dong and Chai, Tianyou},
  journal={IEEE Transactions on Automatic Control},
  volume={69},
  number={4},
  pages={2737--2744},
  year={2023},
  publisher={IEEE}
}

@inproceedings{falanga2017vision,
  title={Vision-based autonomous quadrotor landing on a moving platform},
  author={Falanga, Davide and Zanchettin, Alessio and Simovic, Alessandro and Delmerico, Jeffrey and Scaramuzza, Davide},
  booktitle={2017 IEEE International Symposium on Safety, Security and Rescue Robotics (SSRR)},
  pages={200--207},
  year={2017},
  organization={IEEE}
}

@article{orekhov2023inspiring,
  title={Inspiring field robotics advances through the design of the darpa subterranean challenge},
  author={Orekhov, Viktor L and Maio, Angela C and Daniel, Roshan P and Chung, Timothy H},
  journal={Field Robotics},
  volume={3},
  pages={560--604},
  year={2023},
  publisher={FRPS}
}

@inproceedings{liang2020freebot,
  title={Freebot: A freeform modular self-reconfigurable robot with arbitrary connection point-design and implementation},
  author={Liang, Guanqi and Luo, Haobo and Li, Ming and Qian, Huihuan and Lam, Tin Lun},
  booktitle={2020 IEEE/RSJ International Conference on Intelligent Robots and Systems (IROS)},
  pages={6506--6513},
  year={2020},
  organization={IEEE}
}

@article{daudelin2018integrated,
  title={An integrated system for perception-driven autonomy with modular robots},
  author={Daudelin, Jonathan and Jing, Gangyuan and Tosun, Tarik and Yim, Mark and Kress-Gazit, Hadas and Campbell, Mark},
  journal={Science Robotics},
  volume={3},
  number={23},
  pages={eaat4983},
  year={2018},
  publisher={American Association for the Advancement of Science}
}

@article{xu2021fast,
  title={Fast-lio: A fast, robust lidar-inertial odometry package by tightly-coupled iterated kalman filter},
  author={Xu, Wei and Zhang, Fu},
  journal={IEEE Robotics and Automation Letters},
  volume={6},
  number={2},
  pages={3317--3324},
  year={2021},
  publisher={IEEE}
}

@inproceedings{faessler2014monocular,
  title={A monocular pose estimation system based on infrared leds},
  author={Faessler, Matthias and Mueggler, Elias and Schwabe, Karl and Scaramuzza, Davide},
  booktitle={2014 IEEE international conference on robotics and automation (ICRA)},
  pages={907--913},
  year={2014},
  organization={IEEE}
}

@article{narvaez2020autonomous,

  title={Autonomous VTOL-UAV docking system for heterogeneous multirobot team},
  author={Narv{\'a}ez, Eduardo and Ravankar, Ankit A and Ravankar, Abhijeet and Emaru, Takanori and Kobayashi, Yukinori},
  journal={IEEE Transactions on Instrumentation and Measurement},
  volume={70},
  pages={1--18},
  year={2020},
  publisher={IEEE}
}

@article{oberacker2026mosaic,
  title={MOSAIC: Modular scalable autonomy for intelligent coordination of heterogeneous robotic teams},
  author={Oberacker, David and Richter, Julia and Arm, Philip and Besselmann, Marvin Grosse and Puck, Lennart and Talbot, William and Schik, Maximilian and Bellmann, Sabine and Schnell, Tristan and Kolvenbach, Hendrik and others},
  journal={arXiv preprint arXiv:2601.23038},
  year={2026}
}

@inproceedings{carlson2022armvtol,
  title={A Multi-{VTOL} Modular Aspect Ratio Reconfigurable Aerial Robot},
  author={Carlson, Stephen J. and Arora, Prateek and Papachristos, Christos},
  booktitle={2022 IEEE International Conference on Robotics and Automation (ICRA)},
  pages={8--15},
  year={2022},
  organization={IEEE},
  doi={10.1109/ICRA46639.2022.9811542}
}

@inproceedings{carlson2023multiday,
  title={Towards Multi-Day Field Deployment Autonomy: A Long-Term Self-Sustainable Micro Aerial Vehicle Robot},
  author={Carlson, Stephen J. and Arora, Prateek and Karakurt, Tolga and Moore, Brandon and Papachristos, Christos},
  booktitle={2023 IEEE International Conference on Robotics and Automation (ICRA)},
  pages={11396--11403},
  year={2023},
  organization={IEEE},
  doi={10.1109/ICRA48891.2023.10161014}
}

@inproceedings{depetris2022marsupial,
  title={Marsupial Walking-and-Flying Robotic Deployment for Collaborative Exploration of Unknown Environments},
  author={De Petris, Paolo and Khattak, Shehryar and Dharmadhikari, Mihir and Waibel, Gabriel and Nguyen, Huan and Montenegro, Markus and Khedekar, Nikhil and Alexis, Kostas and Hutter, Marco},
  booktitle={2022 IEEE International Symposium on Safety, Security, and Rescue Robotics (SSRR)},
  pages={188--194},
  year={2022},
  organization={IEEE},
  doi={10.1109/SSRR56537.2022.10018768}
}

@article{deng2023selfspin,
  title={Self-Spin Enabled Docking and Detaching of a {UAV--UGV} System for Aerial-Terrestrial Amphibious and Independent Locomotion},
  author={Deng, Lingxiao and Yang, Binqi and Dong, Xin and Cui, Yangjie and Gao, Yuzhe and Li, Daochun and Tu, Zhan},
  journal={IEEE Robotics and Automation Letters},
  volume={8},
  number={5},
  pages={2454--2461},
  year={2023},
  doi={10.1109/LRA.2023.3254445}
}

@inproceedings{duffy2015lift,
  title={The lift! project-modular, electric vertical lift system with ground power tether},
  author={Duffy, Michael J and Samaritano, Anthony},
  booktitle={33rd AIAA applied aerodynamics conference},
  pages={3013},
  year={2015}
}

@article{oung2011distributed,
  title={The distributed flight array},
  author={Oung, Raymond and D’Andrea, Raffaello},
  journal={Mechatronics},
  volume={21},
  number={6},
  pages={908--917},
  year={2011},
  publisher={Elsevier}
}

@article{richter2026practical,
  title={A Practical Framework of Key Performance Indicators for Multi-Robot Lunar and Planetary Field Tests},
  author={Richter, Julia and Oberacker, David and Ligeza, Gabriela and Bickel, Valentin T and Arm, Philip and Talbot, William and Besselmann, Marvin Grosse and Kehl, Florian and Schnell, Tristan and Kolvenbach, Hendrik and others},
  journal={arXiv preprint arXiv:2601.20529},
  year={2026}
}

@article{tranzatto2022cerberus,
  title={Cerberus in the darpa subterranean challenge},
  author={Tranzatto, Marco and Miki, Takahiro and Dharmadhikari, Mihir and Bernreiter, Lukas and Kulkarni, Mihir and Mascarich, Frank and Andersson, Olov and Khattak, Shehryar and Hutter, Marco and Siegwart, Roland and others},
  journal={Science Robotics},
  volume={7},
  number={66},
  pages={eabp9742},
  year={2022},
  publisher={American Association for the Advancement of Science}
}

@article{lindqvist2022multimodality,
  title={Multimodality Robotic Systems: Integrated Combined Legged-Aerial Mobility for Subterranean Search-and-Rescue},
  author={Lindqvist, Bj{\"o}rn and Karlsson, Samuel and Koval, Anton and Tevetzidis, Ilias and Halu{\v{s}}ka, Jakub and Kanellakis, Christoforos and Agha-Mohammadi, Ali-Akbar and Nikolakopoulos, George},
  journal={Robotics and Autonomous Systems},
  volume={154},
  pages={104134},
  year={2022},
  doi={10.1016/j.robot.2022.104134}
}

@article{liu2023transportation,
  title={Enhancing the Terrain Adaptability of a Multirobot Cooperative Transportation System via Novel Connectors and an Improved Control Framework},
  author={Liu, Quan and Gong, Zhao and Nie, Zhenguo and Liu, Xin-Jun},
  journal={Frontiers of Mechanical Engineering},
  volume={18},
  number={3},
  pages={38},
  year={2023},
  doi={10.1007/s11465-023-0754-2}
}

@inproceedings{saldana2018modquad,
  title={{ModQuad}: The Flying Modular Structure that Self-Assembles in Midair},
  author={Salda{\~n}a, David and Gabrich, Bruno and Li, Guanrui and Yim, Mark and Kumar, Vijay},
  booktitle={2018 IEEE International Conference on Robotics and Automation (ICRA)},
  pages={691--698},
  year={2018},
  organization={IEEE},
  doi={10.1109/ICRA.2018.8461014}
}

@article{sihite2023morphobot,
  title={Multi-Modal Mobility Morphobot ({M4}) with Appendage Repurposing for Locomotion Plasticity Enhancement},
  author={Sihite, Eric and Kalantari, Arash and Nemovi, Reza and Ramezani, Alireza and Gharib, Morteza},
  journal={Nature Communications},
  volume={14},
  pages={3323},
  year={2023},
  doi={10.1038/s41467-023-39018-y}
}

@article{zhu2024takeoff,
  title={Adaptive Take-Off Controller of a Land--Air Amphibious Vehicle on Unstructured Terrain},
  author={Zhu, Hua and Fan, Wei and Xu, Bin and Yang, Chao and Bai, Weiqi and Qin, Yechen and Xu, Tao},
  journal={IEEE Transactions on Vehicular Technology},
  volume={73},
  number={2},
  pages={1817--1828},
  year={2024},
  doi={10.1109/TVT.2023.3319971}
}

@article{liu2023localization,
  author  = {Liu, Yanbaihui},
  title   = {Localization and Navigation System for Indoor Mobile Robot},
  journal = {Highlights in Science, Engineering and Technology},
  volume  = {43},
  pages   = {198--206},
  year    = {2023},
  doi     = {10.54097/hset.v43i.7420}
}

\end{document}